\documentclass[11pt]{article}

\usepackage{scicite}

\usepackage{times}

\usepackage[margin=1.0in]{geometry}

\usepackage{setspace}

\usepackage{multirow}

\usepackage{graphicx}

\newenvironment{sciabstract}{%
\begin{quote} }
{\end{quote}}

\title{Protein Secondary Structure Prediction Using Deep Multi-scale Convolutional Neural Networks and Next-Step Conditioning} 

\author
{Akosua Busia$^{\dagger \ast}$, Jasmine Collins$^{\ast}$, Navdeep Jaitly\\
\\
\normalsize{Google Brain, Mountain View, CA, USA}\\
\normalsize{$^{\dagger}$ Contact author: \texttt{apbusia@google.com}}\\
\\
\normalsize{$^{\ast}$ Work completed as a member of the Google Brain Residency program (g.co/brainresidency)}\\
}

\date{}

\begin{document} 


\baselineskip14pt

\maketitle 

\begin{sciabstract}
Recently developed deep learning techniques have significantly improved the accuracy of various speech and image recognition systems. In this paper we adapt some of these techniques for protein secondary structure prediction.  We first train a series of deep neural networks to predict eight-class secondary structure labels given a protein's amino acid sequence information and find that using recent methods for regularization, such as dropout and weight-norm constraining, leads to measurable gains in accuracy. We then adapt recent convolutional neural network architectures--Inception, ReSNet, and DenseNet with Batch Normalization--to the problem of protein structure prediction. These convolutional architectures make heavy use of multi-scale filter layers that simultaneously compute features on several scales, and use residual connections to prevent underfitting. Using a carefully modified version of these architectures, we achieve state-of-the-art performance of 70.0\% per amino acid accuracy on the public CB513 benchmark dataset. Finally, we explore additions from sequence-to-sequence learning, altering the model to make its predictions conditioned on both the protein's amino acid sequence and its past secondary structure labels. We introduce a new method of ensembling such a conditional model with our convolutional model, an approach which reaches 70.6\% Q8 accuracy on the same test data. We argue that these results can be further refined for larger boosts in prediction accuracy through more sophisticated attempts to control overfitting of conditional models. We aim to release the code for these experiments as part of the TensorFlow repository.
\end{sciabstract}

\section*{Introduction}

Protein-based interactions are responsible for controlling a variety of vital functions: they are critical in driving the immune system, regulating breathing and oxygenation, controlling aging and energy usage, and determining drug response. A protein's particular functional role is determined by its structure.$^{1, 2}$ Over time, scientists have reached a consensus that a protein's structure primarily depends on its amino acid sequence--local and long-range interactions between amino acids and their side-chains are both a cause and consequence of protein secondary and tertiary structure.$^{3}$ This hypothesis has driven a decades-long effort, spanning multiple disciplines, to deduce how protein sequence determines a protein's structure and functional properties.$^{3, 4}$

As the number of proteins with known sequence continues to outpace the number of experimentally determined secondary and tertiary structures,$^{15}$ computational approaches to protein structure prediction become increasingly desirable. Computational tools that can handle large amounts of data while making sufficiently accurate predictions of secondary structures can potentially serve to mitigate the cost and time burden in the experimental determination of three-dimensional structures of proteins.

\section*{Related Works}

Applications of machine learning to the protein structure problem have a rich history. The use of neural networks for secondary structure prediction was pioneered in 1988,$^5$ and subsequent attempts using both recurrent $^{6, 7}$ and convolutional $^{7, 8, 9, 10}$ neural network architectures have since managed to incrementally improve state-of-the-art accuracy on the eight-class secondary structure problem. For example, by combining a deep convolutional neural network with a Conditional Random Field for structure prediction, Wang et al.$^{10}$ demonstrate convincing improvements from the use of consecutive convolutions, achieving 68.3\% per amino acid accuracy on test. In contrast, Li \& Yu$^{7}$ used an ensemble of ten independently trained models, each comprised of a multi-scale convolutional layer followed by three stacked bidirectional recurrent layers, to achieve 69.7\% per amino acid accuracy on the same test set, the highest accuracy previously reported.

The current work improves upon the test accuracy of these models by adapting several techniques developed in the last two years for refining the accuracy of neural networks. We rely substantially on recent general purpose improvements in the field of deep learning--including Batch Normalization, dropout, and weight constraining $^{11, 12}$--as well as architectural innovations, such as ResNET, in the specific domains of image recognition $^{13, 14, 15, 16}$ and sequence-to-sequence learning.$^{17, 18}$ In so doing, we develop a series of new baseline models which demonstrate the impact of these refinements.

We present two main contributions. First, we perform an in-depth analysis of the use of recent convolutional architectures for the eight-class protein secondary structure problem. While such techniques have received significant attention and analysis in image recognition, here we show how to adapt these architectures to protein structure prediction and achieve a new state-of-the-art accuracy. Unlike the prior state-of-the-art model, which uses an ensemble of ten models and multitask learning, our result here is achieved with a single model and could be further improved by applying these techniques. Next, we explore conditioning of future predictions on both sequence data and past structure labels, as done in sequence-to-sequence learning.$^{17, 18}$ While the resulting models currently exhibit significant overfitting, we attain a second state-of-the-art result by introducing a new technique to ensemble these conditional models with convolutional models, achieving 70.6\% Q8 accuracy on test. We expect that these models will lead to further gains in secondary structure prediction as the overfitting problem is more thoroughly addressed.

\section*{Data}

\renewcommand{\thefootnote}{\fnsymbol{footnote}}

We use two publicly available benchmark datasets, preprocessed by Zhou \& Troyanskaya:$^{8}$\footnote{\url{http://www.princeton.edu/~jzthree/datasets/ICML2014/}} CullPDB and CB513. Each of these consists of protein sequences and structure label assignments downloaded from the Protein Data Bank archive (PDB).$^{21}$ Each protein is represented as a sequence of amino acids, padded, if necessary, to a sequence length of 700 residues. In turn, each amino acid is encoded as a 42-dimensional vector: the first twenty-one dimensions represent the one-hot encoding of the amino acid's identity (or a no-sequence padding marker), while the remaining dimensions contain Position-Specific Scoring Matrices (PSSM) generated with PSI-BLAST (see $^{8}$ for more details) which we normalize via mean-centering and scaling by the standard deviation. The full CullPDB dataset contains 6,128 proteins with less than 30\% sequence identity. Here, we use a subset of these data, which has been filtered to reduce sequence identity with the CB513 test data to at most 25\%; this results in a set of 5,534 protein sequences. Consistent with S{\o}nderby \& Winther's${^6}$ arrangement, we randomly divide these 5,534 proteins into a training set of 5,278 proteins and a validation set of 256 proteins. The 513 protein sequences in the CB513 dataset are used exclusively to measure test Q8 accuracy--that is, the percentage of residues for which the predicted secondary structure labels are correct on test.

\section*{Methods}

We use TensorFlow's ADAM optimizer$^{19}$ with default beta and epsilon parameters to train our neural networks. During training, we use a randomly selected mini-batch of 54 proteins on each iteration.

\subsection*{Fully-Connected Model}

We begin with a simplistic baseline to give us a sense of the fundamental difficulty of the eight-class secondary structure problem, as well as the impact of effective use of regularization techniques for deep learning. In particular, we feed a fixed-sized context window of seventeen amino acids (padding edge-cases with no-sequence markers) through five feed-forward layers of 455 rectified-linear units each in order to predict the secondary structure label of the central amino acid via a softmax output layer. We use a learning rate of 0.0004--which is reduced by 50\% every 35,000 training iterations---with the ADAM optimizer,$^{19}$ and perform early-stopping based Q8 accuracy on the validation set. In addition, we use a combination of dropout and max-norm constraining on the model weights; this has been shown to give notable performance improvements even when the number of model parameters drastically outnumbers the size of the dataset.$^{12}$ Previous works on secondary structure prediction have used dropout,$^{6, 7, 9}$ but have not combined dropout with weight constraining. We find significant improvements by randomly dropping out 20\% of the units of each layer and constraining the $l2$-norm of every unit's incoming weights to be at most 0.04614.

This feedforward architecture is able to synthesize local information within a fixed distance of each amino acid, and thus makes the assumption that secondary structure at a particular point in the protein sequence is uniquely determined by information provided by the handful of residues immediately preceding and following it. Although not particularly well-suited to the protein secondary structure problem due to this inability to integrate any sequence information outside of the fixed amount of context, we present results from this simple network to gauge the impact of these dropout and weight constraining techniques on the final accuracy.

\subsection*{Convolutional Models}

Convolutions improve upon a fixed-sized window method of sequence analysis by introducing the ability to learn and maintain information about sequence dependencies at different scales. Specifically, filters in lower layers focus on extracting information from local context, and filters in higher layers cover correlations which are more spatially spread-out in the input sequence. Recently developed architectures such as Inception$^{14}$ take this one step further by using convolutional layers which apply different filter sizes simultaneously at each layer. These multi-scale architectures have led to significant improvements in image recognition over vanilla deep convolutional models.$^{14, 16}$ 

We digress slightly, to underscore a fundamental difference in application of convolutions to protein structures when compared to vision tasks. Protein secondary structure prediction, as currently phrased, only has temporal structure over the residues, and not the spatial structure typical of images. Thus, we use filters that are convolutional in this temporal dimension only. This changes our terminology slightly from use in vision problems: rather than (for example) 3x3 filters, which in a vision task implies a filter which looks at a window of three pixels in the $x$-dimension by three pixels in the $y$-dimension, we will refer to 3-filters, implying a convolutional filter which looks at inputs from three amino acids.

We develop a series of benchmark convolutional models, as described in Table \ref{table:1}. We begin by adding a single convolutional layer to our feed-forward model, and subsequently add more complex convolutional structures. Batch Normalization$^{11}$ after each convolution has been shown to accelerate training on image recognition tasks. We find that, for our problem, introducing batch normalization just prior to the application of rectified-linear non-linearity, along with 40\% dropout, provides a strong boost in validation accuracy. In addition, each model maintains max-norm constraining on the weights of the fully-connected layers. For brevity, we focus here on the details of our final convolutional architecture, which outperforms previously reported attempts at using convolutions for protein secondary structure prediction. $^{7, 8, 9, 10}$

\begin{figure}[ht!]
\centering
\includegraphics[width=0.95\textwidth]{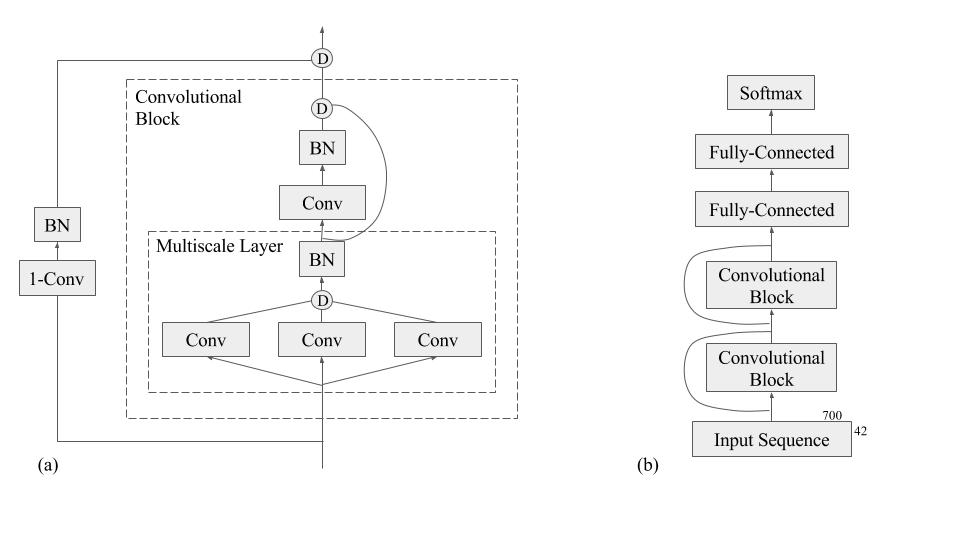}
\caption{(a) Convolutional block composed of consecutive multi- and single-scale convolutions with Batch Normalization and rectified-linear activation (BN), plus depth concatenation (D) of their outputs with modified residual connections between blocks. (b) The final convolutional architecture, composed of two convolutional blocks in the style of (a) followed by two fully-connected layers and a softmax output layer with 40\% dropout between each layer.}
\label{fig:conv}
\end{figure}

Similar to the Inception architecture,$^{14, 16}$ we use blocks composed of single- and multi-scale convolutional layers (see Figure \ref{fig:conv}(a)) as part of the larger neural network displayed in Figure \ref{fig:conv}(b). Table \ref{table:1} describes the number of filters, filter widths, and other architectural adaptions. However, we make further tweaks to the typical Inception network, described below, which improve our results on the protein secondary structure problem.

We use a variant of residual connections introduced in ResNET$^{13}$ that gives better accuracy on predicting protein secondary structure. The ResNET model uses additive identity shortcuts between the outputs of lower layers and the inputs to higher layers to enable more efficient training by improving flow of information throughout the network.$^{16}$ In our models we find application of 1-filter convolutions and depth concatenation of features gives better improvements than the use of typical residual connections. Our choice of depth concatenation is motivated by Huang, Liu, \& Weinberger's$^{15}$ work on the DenseNet architecture. In DenseNet 1x1-filter convolutions are used on the skip connections between blocks to avoid an explosion in the number of features with the addition of blocks to the model--an idea derived from the NiN architecture$^{20}$ which makes use of small 1x1 convolutions to control for the high-dimensionality induced by adding successive multi-scale modules. Thus, rather than using true identity skip connections of ResNET, we use a 1-filter with depth 96 to condense the information learned by the previous block to a smaller, fixed-number of features before concatenating it to the output of the current block. The modified residual connection between blocks formed by these operations is illustrated in Figure \ref{fig:conv}(a).

Our final convolutional model consists of two convolutional blocks connected by the modified residual connections in Figure \ref{fig:conv}(a) followed by two fully-connected layers of 455 units, the first of which receives a fixed-context window of eleven residues (see Figure \ref{fig:conv}(b)). Each multi-scale layer contains 3-, 7-, and 9-filters of depth 64, and is followed by a single convolution with a filter size of 9 and a depth of 24. We use a dropout rate of 40\% on every layer in the network, and apply a max-norm constraint of 0.1503 on the $l2$-norm of the weights of the fully-connected layers. We, again, use the ADAM optimizer$^{19}$ to train the network, this time with a smaller learning rate of 3.357$e^{-4}$ which is reduced by 50\% every 100,000 training iterations. 

\subsection*{Conditioning}

The convolutional model described in the preceding section is motivated by its ability to capture sequence information at multiple scales in the input sequence. However, a convolutional model, as described, does not capture any interdependency between adjacent secondary structure labels because each label is predicted independently of the others. DeepCNF$^{10}$ attempts to integrate information about the correlation between the structures of adjacent residues into a neural network by using a Conditional Random Field (CRF) on the predicted structure labels. In contrast, we pull from sequence-to-sequence learning advances, which have demonstrated improvements by instead accounting for dependencies between output tokens by conditioning the current prediction on the previous labels in addition to the current input.$^{17, 18}$ Unlike CRF, this approach makes fewer assumptions about the independence between the structure labels at different points in the protein sequence, and has the potential to learn richer distributions over our target sequences of secondary structure labels.$^{18}$

More specifically, we attempt to model the secondary structure ${\bf y} = y_1, y_2 \cdots y_L$ of a protein sequence of length $L$, using its input amino acid sequence ${\bf x}=x_1, x_2, \cdots x_L$. Here, $y_i$ is the label for the secondary structure of the amino acid at index $i$, and $x_i$ is the amino acid at index $i$, or more generally, its descriptors, such as real-valued sequence information from PSSM matrices. We attempt to model the probability distribution of $y_1 \cdots y_L$ using a chain rule decomposition:
\begin{equation}
p\left(y_1, \cdots y_L | x_1 \cdots x_L\right) = p\left(y_0 | \bf{x}\right) \prod_{i=2}^L p\left(y_i | \bf{x}, y_{<i}\right)  \nonumber
\end{equation}
This differs from the typical approach taken by convolutional and recurrent models, which optimize a model under the following conditional distribution assumption: 
\begin{equation}
p\left(y_1, \cdots y_L | x_1 \cdots x_L\right) = p\left(y_0 | \bf{x}\right) \prod_{i=2}^L p\left(y_i | \bf{x}\right) \nonumber
\end{equation}

By attempting to learn the distribution $p\left(y_i | \bf{x}, y_{<i}\right)$ we hope to learn about the impact of previous secondary structure labels on the current one. This conditioning on the past labels makes the model learn more general distributions, and avoids conditional independence assumptions amongst the labels of neighboring amino acid residues.

\begin{figure}[ht!]
\centering
\includegraphics[width=0.65\textwidth]{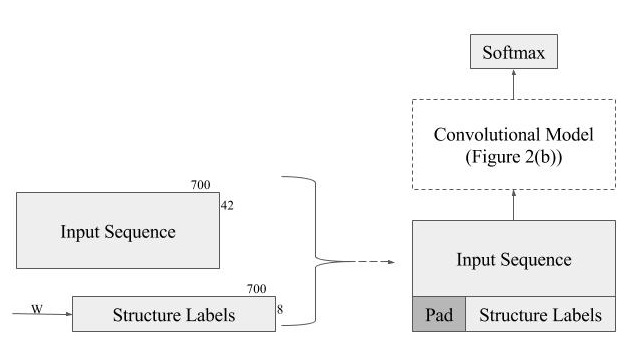}
\caption{Demonstrates the introduction of next-step conditioning on previous labels by including a shifted copy of the input sequence's secondary structure labels as input to the convolutional model from Figure \ref{fig:conv}(b).}
\label{fig:cond}
\end{figure}

In sequence-to-sequence modeling, this conditioning is typically implemented using a combination of recurrent and feedforward layers; here, we instead apply the idea to our convolutional architecture from the previous section. Holding the rest of the architecture constant, we introduce conditioning on a context of previous secondary structure labels during training by appending the input sequence with a copy of the secondary structure labels shifted by the effective total window size $ W$ of the convolutional model. Note that $W$ can be calculated from the network structure; for example for our multiscale architecture, $W = F + n * (C + M - 2)$ where $n$ is the number of convolutional blocks and $F, C, $ and $M$ are the maximum size of the window seen by the fully-connected, convolutional, and multi-scale layers, respectively (see Figure \ref{fig:cond}). This means that, for a given residue in the sequence, the model will integrate information not only from the relevant surrounding amino acids, but also the additional context provided by the previous $W$ secondary structure labels. Thus, we effectively model $p\left(y_i | \bf{x}, y_{<i}\right)$ as $p\left(y_i | \bf{x}, y_{(i-W) \cdots (i-1)}\right)$ using this convolutional neural network.

During inference, we find the label sequence ${\bf y}$ that maximizes $p\left(y_1, \cdots y_L | x_1 \cdots x_L\right)$. To do so, we decode the input sequence using a standard beam search with a beam size of eight over the possible next secondary structure labels; that is, on each step we select the next eight most likely labels assignments based on the running logarithmic probabilities produced by the convolutional model and attempt feeding each of those as additional context to the next step.

We find that the model is able to learn during training to achieve a fairly high accuracy at predicting the next label, if the previous labels are indeed correct; however, during inference the previous labels are not actually correct--rather they are sampled from the predicted distribution over the eight possible structure labels--which hurts the performance of the model. We introduce a simple technique for mitigating some of this overfitting by ensembling this conditioned model with the unconditioned convolutional model. In particular, we combine the predictions of the two models by performing a modified beam search -- at each step in the sequence we select the most likely next structure labels by considering the convex combination of the logarithmic probabilities over the secondary structure labels produced by the unconditional and conditional convolutional architectures. We, again, take a beam size of eight, and use a blending factor of 0.45, so that slightly more weight is given to the predictions made by the unconditional convolutional model from the preceding section.

As a control to examine the impact of the conditioning in this method, we also ensemble two unconditioned convolutional models, weighting their predictions equally, and directly compare the accuracy.

\section*{Results}

With the use of dropout and max-norm constraining, an otherwise vanilla feedforward network is able to perform surprisingly well, achieving 71.4\% Q8 accuracy on validation and 66.8\% on the CB513 dataset. This outperforms SC-GSN's$^{8}$ test metric of 66.4\% and almost matches the 66.9\% Q8 accuracy achieved by Li \& Yu$^{7}$ through a combination of multi-scale convolutions and a forward-passing recurrent neural network, despite being immensely more simplistic. This serves to suggest that a fair amount of the information relevant to predicting an amino acid's secondary structure does, indeed, arise from the local interactions amongst directly neighboring residues. It also serves to highlight the boost in accuracy achieved by subsequent alterations to the model which give it access to information from larger portions of the input sequence.

\begin{table}[ht!]
\centering
\small
\begin{tabular}{|p{1.75cm}|p{1.75cm}|p{1.25cm}|p{1.6cm}|p{1.6cm}|p{1.8cm}|p{1.35cm}|p{1.35cm}|}
\hline
Multi-scale Convolution & Single \ \ \ Convolution & \# Blocks & Fully-Connected Window & Fully-Connected Layers & Residual Connections & Validation Accuracy & CB513 Accuracy \\
\hline\hline
- & - & - & 17 & 5 & N & 71.4\% & 66.8\% \\
\hline
- & 7 x 32 & 1 & 17 & 5 & N & 71.8\% & 67.2\% \\
\hline
- & 7 x 32 & 2 & 17 & 5 & N & 72.2\% & 67.4\% \\
\hline
3 x 32 & \multirow{3}{4em}{ - } & \multirow{3}{4em}{1} & \multirow{3}{4em}{11} & \multirow{3}{4em}{5} & \multirow{3}{4em}{N} & \multirow{3}{4em}{72.5\%} & \multirow{3}{4em}{67.4\%} \\
5 x 32 & & & & & & & \\
7 x 32 & & & & & & & \\
\hline
3 x 32 & \multirow{3}{4em}{ - } & \multirow{3}{4em}{1} & \multirow{3}{4em}{11} & \multirow{3}{4em}{2} & \multirow{3}{4em}{N} & \multirow{3}{4em}{72.8\%} & \multirow{3}{4em}{68.1\%} \\
5 x 32 & & & & & & & \\
7 x 32 & & & & & & & \\
\hline
3 x 32 & \multirow{3}{4em}{7 x 32} & \multirow{3}{4em}{1} & \multirow{3}{4em}{11} & \multirow{3}{4em}{2} & \multirow{3}{4em}{N} & \multirow{3}{4em}{72.8\%} & \multirow{3}{4em}{68.2\%} \\
5 x 32 & & & & & & & \\
7 x 32 & & & & & & & \\
\hline
3 x 64 & \multirow{3}{4em}{9 x 24} & \multirow{3}{4em}{2} & \multirow{3}{4em}{11} & \multirow{3}{4em}{2} & \multirow{3}{4em}{N} & \multirow{3}{4em}{74.3\%} & \multirow{3}{4em}{69.0\%} \\
7 x 64 & & & & & & & \\
9 x 64 & & & & & & & \\
\hline
3 x 64 & \multirow{3}{4em}{9 x 24} & \multirow{3}{4em}{5} & \multirow{3}{4em}{11} & \multirow{3}{4em}{2} & \multirow{3}{4em}{N} & \multirow{3}{4em}{73.8\%} & \multirow{3}{4em}{69.1\%} \\
7 x 64 & & & & & & & \\
9 x 64 & & & & & & & \\
\hline
3 x 64 & \multirow{3}{4em}{9 x 24} & \multirow{3}{4em}{2} & \multirow{3}{4em}{11} & \multirow{3}{4em}{2} & \multirow{3}{4em}{Y} & \multirow{3}{4em}{74.8\%} & \multirow{3}{4em}{70.0\%} \\
7 x 64 & & & & & & & \\
9 x 64 & & & & & & & \\
\hline
\end{tabular}
\caption{Displays architectural details as well as validation and test accuracies for the series of baseline convolutional models.}
\label{table:1}
\end{table}

Our incremental study of convolutional architectures (see Table \ref{table:1}) highlights that introduction of convolutions improves accuracy--the addition of merely a single convolutional filter prior to the feedforward model improves Q8 accuracy on CB513 by roughly 0.4\%. In addition, we demonstrate a benefit from multi-scale convolutions versus single convolutional filters, finding that one multi-scale layer prior to five fully-connected layers results in comparable accuracy to two convolutional layers before the same feedforward network, despite the wider overall context window seen by the latter. In addition, we find that the final few fully-connected layers become unnecessary with the addition of multi-scale layers, with a reduction to two such layers being slightly preferable to the full five. However, a combination of single and multi-scale convolutional layers greatly outperforms either, increasing test accuracy by roughly 0.8\% regardless of a decrease in the size of the context window into the final two layers. Finally, we examine the impact of our adapted residual connections, finding that introducing these connection results in a much larger accuracy improvement than the 0.1\% increase from simply adding more convolutional blocks to the model. We believe these connections improve the model's ability to retain information about smaller local contexts within the sequence, without them being washed out by longer-range features learned by subsequent convolutions over larger patches of the input sequence. This addition brings the Q8 accuracy of our final convolutional architecture on the CB513 test dataset to 70.0\%. This improves upon the previous state of the art by roughly 0.3\%, despite their use of model ensembling, recurrence, and multitask learning to achieve additional accuracy gains (see Table \ref{table:2}).

\begin{table}[ht!]
\centering
\small
\begin{tabular}{|c|c|c|c|}
\hline
Model & Multitasking & Ensembling & CB513 Accuracy\\
\hline\hline
SC-GSN$^{8}$ & Y & N & 66.4\% \\
Our Feedforward & N & N & 66.8\% \\
Multi-scale Conv + GRU$^{7}$ & Y & N & 66.9\% \\
bLSTM$^{6}$ & N & N & 67.4\% \\
DeepCNF$^{10}$ & N & N & 68.3\% \\
MUST-CNN$^{9}$ & Y & N & 68.4\% \\
Multi-scale Conv + bGRU$^{7}$ & Y & Y & 69.7\% \\
Our CNN & N & N & 70.0\% \\
Our CNN + Conditioned CNN Ensemble & N & Y & 70.6\% \\
\hline
\end{tabular}
\caption{Places the results of the current work in context of other recent published results applying deep learning to the eight-class secondary structure problem.}
\label{table:2}
\end{table}

We find that adding conditioning on past secondary structure labels to the convolutional architecture has a large impact. This model achieves a next-step Q8 accuracy of 81.7\% on the validation set and 77.0\% on CB513 when the ground truth secondary structure labels are fed as past context. However, when inference is applied with beam search, the validation and test Q8 accuracies reduce to 71.9\% and 67.1\%, respectively. We note that this is lower than the 68.3\% Q8 accuracy achieved on the CB513 dataset by DeepCNF.$^{10}$ This suggests that such an unstructured approach to conditioning on the past induces significant overfitting for protein secondary structure prediction; it is likely that the combination of the lack of assumptions regarding the nature of dependence between structure labels and the repetitive nature of secondary structure sequences causes the model to ``over-learn'' dependencies between consecutive labels. We posit that the model learns to almost always simply copy the previously seen label assignment during training, making it perform well on next-step prediction with ground truth labels, but relatively poorly on actual inference when the sequence has not been seen before and the truth is not known.

We aim to mitigate the negative impact of this copying effect during inference by combining the conditional model with the predictions of our state-of-the-art convolutional model through a weighted beam search. Putting a slightly larger weight on the predictions of the unconditional model injects some of its good predictions into the log probabilities calculated on each step of the search, theoretically reducing the long-term impact of mistakes despite the conditional model's tendency to largely copy previous structure labels. We find that this ensemble results in better accuracy than conditioning alone, resulting in 75.6\% and 70.6\% Q8 accuracy on validation and test. For comparison, ensembling two unconditional convolutional models yields validation and test accuracies of 75.1\% and 70.4\%. This suggests that, while much of the accuracy boost arises simply from ensembling two models, there is some additional benefit to conditioning on the previous $W$ secondary structure predictions. We surmise that this benefit will be much more impactful when the overfitting of the conditional model itself is more adequately controlled.

\section*{Conclusion}

The current work contributes two state-of-the-art results to the eight-class secondary structure prediction problem. By analyzing the impact of convolutional variants developed for recognition of natural images, we develop a modified multi-scale and residual convolutional architecture which outperforms previous deep learning approaches on the same benchmark dataset. In addition, we introduce the idea of using conditioning on past structure labels to boost accuracy. While there is much future work to be done to exploit conditioning by mitigating the overfitting, we demonstrate a boost from ensembling the conditional model with our convolutional model, pushing state-of-the-art Q8 accuracy on the CB513 data further by 0.9\% relative to the highest previously reported result. In future work, we hope to apply these techniques to training on and predicting protein backbone angles instead of coarse secondary structure labels; we anticipate that this approach will suffer less from overfitting induced by conditioning, since exact copying of consecutive output values will no longer yield good accuracy during training.

\singlespace
\bigskip

{\bf References}

\begin{enumerate}
    \item Breda A, Valadares NF, Norberto de Souza O, et al. Protein Structure, Modelling and Applications. 2006 May 1 [Updated 2007 Sep 14]. In: Gruber A, Durham AM, Huynh C, et al., editors. Bioinformatics in Tropical Disease Research: A Practical and Case-Study Approach [Internet]. Bethesda (MD): National Center for Biotechnology Information (US); 2008. Chapter A06. Available from: http://www.ncbi.nlm.nih.gov/books/NBK6824/
    \item Guo, J., Ellrott, K., \& Xu, Y. (2007). A Historical Perspective of Template-Based Protein Structure Prediction. In M. J. Zaki \& C. Bystroff (Eds.), \textit{Protein Structure Prediction} (2nd ed., Vol. 413, Springer Science \& Business Media., pp. 3-31). Springer.
    \item Dill, K. A., Ozkan, S. B., Shell, M. S., \& Weikl, T. R. (2008). The protein folding problem. \textit{Annual review of biophysics}, 37, 289.
    \item Anfinsen, C. B. (1977). Studies on the principles that govern the folding of protein chains. \textit{Nobel Lectures in Molecular Biology}: 1933-1975, 401.
    \item Qian, N., \& Sejnowski, T. J. (1988). Predicting the secondary structure of globular proteins using neural network models. \textit{Journal of molecular biology, 202}(4), 865-884.
    \item S{\o}nderby, S. K., \& Winther, O. (2014). Protein secondary structure prediction with long short term memory networks. \textit{arXiv preprint arXiv:1412.7828.}
    \item Li, Z., \& Yu, Y. (2016). Protein Secondary Structure Prediction Using Cascaded Convolutional and Recurrent Neural Networks. \textit{arXiv preprint arXiv:1604.07176.}
    \item Zhou, J., \& Troyanskaya, O. G. (2014, March). Deep Supervised and Convolutional Generative Stochastic Network for Protein Secondary Structure Prediction. In \textit{ICML} (pp. 745-753).
    \item Lin, Z., Lanchantin, J., \& Qi, Y. (2016). MUST-CNN: A Multilayer Shift-and-Stitch Deep Convolutional Architecture for Sequence-based Protein Structure Prediction. \textit{arXiv preprint arXiv:1605.03004.}
    \item Wang, S., Peng, J., Ma, J., \& Xu, J. (2016). Protein secondary structure prediction using deep convolutional neural fields. \textit{Scientific reports, 6.}
    \item Ioffe, S., \& Szegedy, C. (2015). Batch normalization: Accelerating deep network training by reducing internal covariate shift. \textit{arXiv preprint arXiv:1502.03167.}
    \item Srivastava, N., Hinton, G. E., Krizhevsky, A., Sutskever, I., \& Salakhutdinov, R. (2014). Dropout: a simple way to prevent neural networks from overfitting. \textit{Journal of Machine Learning Research, 15}(1), 1929-1958.
    \item He, K., Zhang, X., Ren, S., \& Sun, J. (2015). Deep residual learning for image recognition. \textit{arXiv preprint arXiv:1512.03385.}
    \item Szegedy, C., Liu, W., Jia, Y., Sermanet, P., Reed, S., Anguelov, D., ... \& Rabinovich, A. (2015). Going deeper with convolutions. In \textit{Proceedings of the IEEE Conference on Computer Vision and Pattern Recognition} (pp. 1-9).
    \item Huang, G., Liu, Z., \& Weinberger, K. Q. (2016). Densely Connected Convolutional Networks. \textit{arXiv preprint arXiv:1608.06993.}
    \item Szegedy, C., Ioffe, S., \& Vanhoucke, V. (2016). Inception-v4, inception-resnet and the impact of residual connections on learning. \textit{arXiv preprint arXiv:1602.07261.}
    \item Sutskever, I., Vinyals, O., \& Le, Q. V. (2014). Sequence to sequence learning with neural networks. In \textit{Advances in Neural Information Processing Systems} (pp. 3104-3112).
    \item Bengio, S., Vinyals, O., Jaitly, N., \& Shazeer, N. (2015). Scheduled sampling for sequence prediction with recurrent neural networks. In \textit{Advances in Neural Information Processing Systems} (pp. 1171-1179).
    \item Kingma, D., \& Ba, J. (2014). Adam: A method for stochastic optimization. \textit{arXiv preprint arXiv:1412.6980.}
    \item Lin, M., Chen, Q., \& Yan, S. (2013). Network in Network. \textit{arXiv preprint arXiv:1312.4400.}
    \item Berman, H.M., Henrick, K., \& Nakamura, H. (2003). Announcing the worldwide Protein Data Bank. \textit{Nature Structural Biology 10}(12): 980.
\end{enumerate}

\end{document}